\documentclass{article}

\usepackage[preprint]{neurips_2024}

\usepackage[utf8]{inputenc} 
\usepackage[T1]{fontenc}    
\usepackage{hyperref}       
\usepackage{url}            
\usepackage{booktabs}       
\usepackage{amsfonts}       
\usepackage{nicefrac}       
\usepackage{microtype}      
\usepackage{xcolor}         
\usepackage{latexsym}
\usepackage{amssymb}
\usepackage{amsmath}
\usepackage{amsthm}
\usepackage{enumerate}
\usepackage{graphicx}
\usepackage{color}
\usepackage{makecell}
\usepackage{bbm}
\usepackage{booktabs, nicematrix}
\usepackage{subcaption}
\usepackage{algorithm}
\usepackage{algorithmic}
\usepackage{xcolor}
\usepackage{bm}
\usepackage[defblank]{paralist}
\usepackage{tikz}
\usetikzlibrary{arrows,matrix, positioning, shapes.geometric}
\usepackage{multirow}

\newtheorem{theorem}{Theorem}

\newtheorem{proposition}[theorem]{Proposition}

\usepackage{amsmath,amsfonts,amssymb,bm}
\usepackage{hyperref}
\usepackage{url}
\usepackage{tikz}
\usetikzlibrary{arrows,matrix, positioning,  shapes.geometric}

\usepackage{algorithm, algorithmic, cancel}
\usepackage{makecell}
\usepackage{pifont}
\usepackage{pgfplots}
\pgfplotsset{compat=1.18}
\usepackage{listofitems} 
\tikzstyle{mynode}=[thick,inner sep={.75\pgflinewidth}, draw=black,fill=white,circle,minimum size=3]

\newcommand\m[1]{\ensuremath{\mathcal{#1}}}
\newcommand{\plan}{\bm{\pi}}
\newcommand{\optplan}[1]{\plan^\star (#1)}

\newcommand{\throptplan}[1]{\plan^\star_\mathit{relu} (#1)}
\newcommand{\addminoptplan}[1]{\plan^\star_{\mathit{min}^+ }(#1)}
\newcommand{\cost}{\bm{C}}
\newcommand{\predcost}{\hat{\cost}}
\newcommand{\negativeoptplan}[1]{ \check{\plan}^\star (#1)}
\newcommand\Regret{\m{{\mathbf{regret}}}}
\newcommand\UpperRegret{\overline{\m{{\mathit{regret}}}}}
\newcommand{\relu}[1]{\m{{\mathit{relu}}} (#1) }
\newcommand\spoplus{\m{{\mathit{SPO}_+}}}

\newcommand\gfunc{\mathit{g}}
\newcommand\nspo{\spoplus_{\mathcal{P}}}
\newcommand\ML{\mathcal{M}_{\bm{\omega}}}
\newcommand\features{\bm{\m{X}}}

\newcommand{\marco}[1]{{\color{red}Marco: #1}}

\newcommand{\jay}[1]{{\color{blue}Jay: #1}}

\newcommand{\rev}[1]{{\color{violet}#1}}

\newcommand{\stchange}[1]{{\color{orange} #1}}
\mathchardef\mhyphen="2D

\newcommand{\plVars}[0]{\ensuremath{P}}
\newcommand{\plActions}[0]{\ensuremath{A}}
\newcommand{\plActionSeq}[0]{\ensuremath{\overline{A}}}
\newcommand{\plInit}[0]{\ensuremath{s_0}}
\newcommand{\plGoal}[0]{\ensuremath{g}}

\newcommand{\plPrecN}[0]{\ensuremath{\mathit{prec}}}
\newcommand{\plAddN}[0]{\ensuremath{\mathit{add}}}
\newcommand{\plDelN}[0]{\ensuremath{\mathit{del}}}
\newcommand{\plCostsN}[0]{\ensuremath{c}}

\newcommand{\plPrec}[1]{\ensuremath{\mathit{\plPrecN(#1)}}}
\newcommand{\plAdd}[1]{\ensuremath{\mathit{\plAddN(#1)}}}
\newcommand{\plDel}[1]{\ensuremath{\mathit{\plDelN(#1)}}}

\newcommand{\plApply}[2]{\ensuremath{\mathit{\gamma(#1, #2)}}}
\newcommand{\sol}{action count vector}
\newcommand{\sols}{action count vectors}

\newcommand{\solOpt}[0]{\ensuremath{\mathit{opt}}}
\newcommand{\solBound}[1]{\ensuremath{\mathit{bound_{#1}}}}
\newcommand{\solNonOpt}[0]{\ensuremath{\mathit{no\mhyphen bound}}}
\newcommand{\solHeuOpt}[0]{\ensuremath{\mathit{h}}}

\newcommand{\hff}[0]{\ensuremath{\mathit{h^{FF}}}}

\title{Decision-Focused Learning to Predict Action Costs for Planning}

\author{%
  Jayanta Mandi\\
  Department of Computer Science\\
  KU Leuven, Belgium\\
  \texttt{jayanta.mandi@kuleuven.be} \\
  \And
  Marco Foschini\\
  Department of Computer Science\\
  KU Leuven, Belgium\\
  \texttt{marco.foschini@kuleuven.be} \\
  \And
  Daniel H{\"o}ller\\
  Saarland Informatics Campus\\
  Saarland University, Germany\\
  \texttt{hoeller@cs.uni-saarland.de}
  \And
  Sylvie  Thi\'ebaux\\
  LAAS-CNRS, 
  Université de Toulouse, France\\
  Australian National University, Australia\\
  \texttt{sylvie.thiebaux@laas.fr}
  \And
  J{\"o}rg Hoffmann \\
  Saarland Informatics Campus\\
  Saarland University, Germany\\
  German Research Center for Artificial Intelligence (DFKI)\\
  \texttt{hoffmann@cs.uni-saarland.de}
  \And
  Tias Guns \\
  Department of Computer Science\\
  KU Leuven, Belgium\\
  \texttt{tias.guns@kuleuven.be} \\
}

\newsavebox{\mybox}
\sbox{\mybox}{%
	\begin{tikzpicture}[x=0.35cm,y=0.5cm]
		\readlist\Nnod{3,4,4,1} 
		\foreachitem \N \in \Nnod{ 
			\foreach \i [evaluate={\x=\Ncnt; \y=\N/2-\i+0.5; \prev=int(\Ncnt-1);}] in {1,...,\N}{ 
				\node[mynode] (N\Ncnt-\i) at (\x,\y) {};
				\ifnum\Ncnt>1 
				\foreach \j in {1,...,\Nnod[\prev]}{ 
					\draw[thick] (N\prev-\j) -- (N\Ncnt-\i); 
				}
				\fi 
			}
		}
	\end{tikzpicture}
}
\begin{document}

\maketitle
\begin{abstract}
In many automated planning applications, action costs can be hard to specify. An example is the time needed to travel through a certain road segment, which depends on many factors, such as the current weather conditions. A natural way to address this issue is to learn to predict these parameters based on input features (e.g., weather forecasts) and use the predicted action costs in automated planning afterward. Decision-Focused Learning (DFL) has been successful in learning to predict the parameters of combinatorial optimization problems in a way that optimizes solution quality rather than prediction quality. This approach yields better results than treating prediction and optimization as separate tasks.
In this paper, we investigate for the first time the challenges of
implementing DFL for automated planning in order to learn to predict
the action costs.
There are two main challenges to overcome:
~(1) planning systems are called during gradient descent learning, to solve planning problems with negative action costs, which are not supported in planning. 
We propose novel
methods for gradient computation to avoid this issue.
~(2) DFL requires repeated planner calls during training, which can
limit the scalability of the method. We experiment with different
methods approximating the optimal plan as well as an easy-to-implement caching mechanism to
speed up the learning process.
As the first work that addresses DFL for automated planning, we
demonstrate that the proposed gradient computation consistently yields
significantly better plans than predictions aimed at minimizing
prediction error; and that caching can temper the computation requirements.

\end{abstract}



\section{Introduction}
%
%
Automated planning generates plans aimed at achieving specific goals in a given environment.
However, in real-world environments some information is hard to access and to specify directly in a model, e.g., in a transportation logistics planning domain \cite{Helmert2014OnTC}, to find route-cost optimal solution requires access to travel time between cities.
In today's world, features that are correlated with these unknown parameters are often available and must be leveraged for enhanced planning.
For instance, travel time depends on various environmental and contextual factors like time-of-day, expected weather conditions (e.g., temperature, precipitation).
Machine learning (ML) can facilitate the prediction of these parameters from those correlated features, which can then be used as parameters in a planning model. 
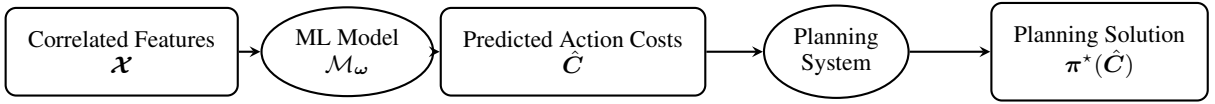
\begin{figure*}
    \centering
    \begin{tikzpicture}[
  font=\rmfamily\footnotesize,
  every matrix/.style={ampersand replacement=\&,column sep=2cm,row sep=.6cm},
  source/.style={draw,thick,rounded corners,inner sep=.3cm},
  process/.style={draw,thick,ellipse, minimum width=1pt,
    align=center},
  dots/.style={gray,scale=2},
  to/.style={->,>=stealth',shorten >=1pt,semithick,font=\rmfamily\scriptsize},
  every node/.style={align=center}, 
  arrow/.style = {thick,-stealth}]

    \node[source] (z) at (-4, 0) {Correlated Features \\ $\features$}; 
     \node[process](ML) at (-1,0){ML Model\\$\ML$ }; 
     \node[source] (param) at (2, 0) {Predicted Action Costs \\ $\predcost$};
     \node[process](Opt) at (5.5,0){Planning\\System}; 
     \node[source](decision) at (9,0){Planning Solution \\ $\optplan{\predcost}$ }; 
     \draw [arrow] (z) --(ML);
     \draw [arrow] (ML) --(param);
     \draw [arrow] (param) --(Opt);
     \draw [arrow] (Opt) --(decision);
\end{tikzpicture}
    \caption{ Predict-then-optimize problem formulation for planning problems.    }
    \label{fig:problem_desc}
\end{figure*}

These two steps, (1) predicting and (2) planning, can in principle be considered as two distinct tasks. 
For example, the approach by \citet{Weiss2023PlanningWM} involves generating a planning solution based on estimates provided by an external model, thus treating prediction and planning as separate tasks.
If the prediction in step (1) is perfect, this would lead to optimal planning in step (2).
However, ML predictions are not always fully accurate, 
for various reasons such as high uncertainty, limited features or ML model capacity, and the presence of noisy data~\cite{hullermeier2021}.

Recent works in \emph{decision-focused learning} (DFL)~\cite{mandi2023decisionfocused} for combinatorial optimization problems has shown that training the ML model to directly optimize the outcome of the downstream optimization problem, rather than prediction quality, leads to higher quality solutions.
Training this way allows the ML  model to focus on parts of the prediction that have (higher) impact on the actual solutions. This has been shown on various combinatorial optimization problems, e.g., shortest path \cite{spo}, knapsack~\cite{MandiDSG20}, TSP~\cite{PogancicPMMR20}, portfolio optimization~\cite{MIPAAL} or energy-cost aware scheduling~\cite{MandiDSG20}.
%

Our interest lies in exploring whether DFL techniques can also be  leveraged to produce plans of higher quality, when having to predict action costs for automated planning.
To the best of our knowledge, this is the first paper on DFL for contextual action cost prediction.

Our starting point is the seminal `Smart Predict-then-Optimize' (SPO) \cite{spo} work for predicting the coefficients of the linear objective function of a combinatorial optimization problem. We will show how this framework is applicable to planning by considering total plan cost as a weighted sum over the action counts of a plan. However, two more fundamental challenges present themselves and form the core of this paper:
First, when using a machine learning system to predict action costs, one might get negative predictions, especially during training. However, for highly non-linear prediction problems this can even be the case when all of the training data has positive values. We hence propose and evaluate two ways of correcting negative predictions. During training, we additionally propose and evaluate an explicit penalty on negative values to guide the learning.

A second challenge is the computational cost of solving planning problems. 
In DFL, we need to call the planner for every training instance, so even with only 100 training instances you easily run into thousand planner calls and more. We hence investigate techniques from planning ~\cite{HoffmannN01} to compute sub-optimal plans and relaxed plans. 
We also experiment with a solution caching approach during learning~\cite{MulambaMD0BG21}, as proposed in DFL for optimization problems.

%
We will empirically demonstrate that the proposed approach generates superior planning solutions compared to predictions aimed solely at minimizing the mean square error (MSE) of action costs.
We also observe that solution caching significantly reduces training time compared to repeatedly solving the optimization problem.
%
%
\section{Background}



We use the STRIPS formalism~\cite{FikesN71} and define a planning problem as a tuple $\left(\plVars, \plActions, \plInit, \plGoal, \plCostsN\right)$, where
\begin{inparablank}
    \item \plVars{} is a set of propositions,
    \item \plActions{} is a set of actions,
    \item $\plInit{} \subseteq \plVars$ is the initial state, 
    \item $\plGoal{} \subseteq \plVars$ the goal definition, and
    \item $\plCostsN: \plActions \rightarrow \mathbb{R}_0$ is the cost function mapping each action to its (positive) costs.
\end{inparablank}
%
%
We identify a state $s \subseteq \plVars$ with the set of propositions that hold in it; propositions that are not included in $s$ are assumed to be \emph{false}. A state $s$ is a \emph{goal state} if and only if $s \subseteq \plGoal$.
%
%
The functions \plPrecN{}, \plAddN{}, and \plDelN{} define the precondition, add-, and delete-effects of actions. Formally, they map actions to subsets of propositions: $\{\plPrecN{}, \plAddN{}, \plDelN{}\} : \plActions \rightarrow 2^{\plVars}$. An action $a$ is applicable in a state $s$ if and only if $\plPrec{a} \subseteq s$. When an applicable action $a$ is applied to a state $s$, the resulting state $s'$ is defined as $s' = \plApply{a}{s} =$ $\left(s \setminus \plDel{a}\right)$ $\cup$ $\plAdd{a}$.

A sequence $(a_0, a_1, \dots, a_n)$ of actions is a solution (or plan) for the problem if and only if each action $a_i$ is applicable in the state $s_i$, with for $i>0$, $s_i = \plApply{a_{i-1}}{s_{i-1}}$, and $s_{n+1} \subseteq g$.
%
%
%
The cost of a solution is defined as the sum of the costs of its actions. By abuse of notation, we define the function $\plCostsN$ also on solutions. Formally, let $p = (a_0, a_1, \dots, a_n)$ be a solution, then $c(p) = \sum_{0 \leq i \leq  n} c(a_i)$. We call a solution $p^\star$
\emph{optimal} with respect to a planning problem when there is no solution $p$ with $c(p^\star) < c(p)$.
Note that there might be multiple optimal solutions to a planning problem.
In our experiments we will assume that if there exist non-unique solutions, the planner returns a single optimal solution by breaking ties in a consistent pre-specified manner. 

\subsection{From Planning to Learning}
\begin{figure}[t]
 \centering
 \includegraphics[width=0.35\textwidth]{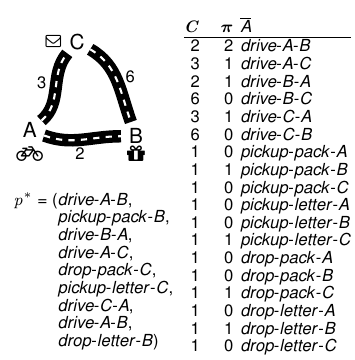}
 \caption{Top left: Illustration of a planning problem. The bike needs to deliver the letter to position $B$ and the package to $C$. It cannot carry both at the same time. The road segments come with different costs, \emph{pickup} and \emph{drop} actions cost $1$. An optimal solution $p^\star$ is given below. 
The right-hand side of the picture successively shows the action costs ($\cost$), the action count vector ($\optplan{\cost}$), and \plActionSeq. 
%
}
 \label{fig:backgroundExample}
\end{figure}

%
\paragraph{Predict-then-Optimize problem.} 
In domains like travelling or delivery services, the costs of actions are hard to specify at design time, because they depend on the current situation, e.g., regarding weather or traffic.
However, one can estimate these costs using contextual features that are correlated with the costs.
In this case, predicting the costs using ML methods is a natural choice. 
When the ground truth action costs are unknown, we employ a trained ML model $\ML$ to predict the action costs from features $\features$. 
The trainable parameters, denoted as $\bm{\omega}$, are estimated using a set of past observations, used as a training dataset for the ML model.
To obtain a feasible plan in this setting, the action costs are first predicted using ML, followed by the generation of a plan optimized with respect to the predicted costs.
This pipeline is commonly referred to as the \emph{Predict-then-Optimize} problem formulation in the literature \citep{spo,MandiDSG20}.
We present a schematic diagram illustrating Predict-then-Optimize in the context of planning problems in Figure~\ref{fig:problem_desc}.
\paragraph{Vector representation.}
%
State-of-the-art ML architectures, including neural networks, represent the data in matrix and vector form. As we will be using neural networks as the predictive model, we will introduce a vector based notation of the action costs and the solution.
Consider the left side of Figure~\ref{fig:backgroundExample}. It shows the illustration of a simple planning problem and an optimal solution as defined before.
%

%
%
We create a vector representation of a plan by storing the 
number of times each action occurs in this plan. Since this discards the orderings of the actions in the plan, more than one plan might map to the same vector.
We refer to this as the \sol\ by $\plan$.
%
More formally, let $m = |\plActions|$ be the number of possible actions $a_i$ in the model, and $\plActionSeq = (a_0, a_1, \dots, a_{m-1})$ a \textit{sequence} containing the actions of $A$ in an arbitrary but fixed ordering.
%
%
Given some plan $p = (p_0, \dots, p_n)$, we define the action count vector
$\plan = (o_0, \dots, o_{m-1})$ with \mbox{$o_i = \sum_{j=0}^{n}\mathbbm{1}(a_i = p_j)$}.

We need a similar vector representation for action costs, which we denote by $\cost$. We define $\cost= (c(a_0), c (a_1), \ldots, c (a_{m-1}))$, where $a_i$ is the $i^{th}$ element of $\plActionSeq$.
%
Hereafter, we will use $\optplan{\cost}$ to denote the \sol\ of an optimal plan 
with respect to $\cost$. 
Observe that both vectors $\cost$ and $\optplan{\cost}$ have the same length. 
%
An example of $\cost$, $\optplan{\cost}$ and $\plActionSeq$ is given on the right of Figure~\ref{fig:backgroundExample}. 
%
With this notation, we can represent the training dataset as $\{ (\features_{\kappa}, \cost_{\kappa}) \}_{\kappa=1}^N$.
\paragraph{Regret.}
In the predict-then-optimize problem, 
we need to distinguish between the ground truth action costs that we want to learn and the predicted action costs. We will denote them as $\cost$ and $\predcost$, respectively.
%
%
%
Let $\optplan{\cost}$ and $\optplan{\predcost}$ be optimal \sols\  with respect to $\cost$ and $\predcost$ respectively.
Using vector notation, 
the cost of executing $\optplan{\cost}$ can be expressed as $ \cost ^\top \optplan{\cost}$.
Importantly, when a plan that was created using the predicted costs is actually executed in practice, the actual costs, $\cost$, is revealed, and the efficacy of the plan is evaluated with respect to $\cost$.
%
Hence the \emph{real} cost of executing $\optplan{\predcost}$ is $\cost^\top \optplan{\predcost}$, e.g. the real cost times the action count vector.
The quality of a predicted cost in a predict-then-optimize problem is evaluated based on \emph{regret}. Regret measures the difference between the realized cost of the solution, made using the predicted cost and the true optimal cost, which is obviously \emph{not} known a priori.
It can be expressed in the following form:
\begin{equation}
\label{eq:regret_def}
\Regret (\predcost, \cost) = \cost^\top \optplan{\predcost} - \cost^\top \optplan{\cost} 
\end{equation}

\subsection{Decision-Focused Learning}
In a predict-then-optimize setup,
the final goal of predicting the cost is to make a planning solution with zero or low regret.
The motivation of DFL is to directly train an ML model to predict
$\predcost$ in a manner that minimizes regret. 
%
We are particularly interested in gradient descent training, a widely utilized method for training neural networks. 
In gradient descent training, the neural network is trained by computing the gradient of the loss function. 
Modern neural network frameworks like TensorFlow~\cite{tensforflow} and PyTorch~\cite{paszke2017automatic} compute this gradient automatically by representing the set of all neural network layers as a \emph{computational graph} \cite{baydin2018automatic}.
However, in DFL, as the final loss is the regret; this 
would require computing the derivative the regret and hence of plan $\optplan{\predcost}$ with respect to $\predcost$.
Firstly, one cannot rely on automatic differentiation to compute this derivative as 
the planning problem is solved outside the neural network computational graph.
Moreover, the planning process is not a differentiable operation, since slight changes in action costs either do not affect the solution or change the solution abruptly, just like in combinatorial optimization~\cite{mandi2023decisionfocused}.
So, the derivative of the planning solution is either zero or undefined.

To obtain gradients useful for DFL for different classes of optimization problems, numerous techniques have been proposed. 
For an extensive analysis of existing DFL techniques, we refer readers to the survey by \citet{mandi2023decisionfocused}.
In this work, we focus on the seminal and robust `Smart Predict then Optimize' (SPO) technique \cite{spo}, which has demonstrated success in implementing DFL across various applications, including e.g.\ power systems \cite{chen2021feature} or antenna design~\cite{chai2022port}.

\paragraph{Smart Predict-then-Optimize (SPO).} 
SPO is a DFL approach which proposes a convex upper-bound of the regret. This upper-bound can be expressed in the following form, as shown below:
\begin{equation}
    \spoplus = \zeta ( \cost -2\predcost) + 2 \predcost^\top  \optplan{\cost} - \cost^\top \optplan{\cost}
\end{equation}
\noindent where 
$\zeta (\cost) \doteq \max_{\plan} \{ \cost^\top \plan \}  $.
However, to minimize the $\spoplus$ loss in gradient-based training, a difficulty arises because it does not have a gradient.
It is easy to verify  that $-\optplan{  \cost}$ is a subgradient of $\zeta (-\cost)$, allowing
to write the following subgradient of $\spoplus$ loss
\begin{equation}
\label{eq:spo_grad}
	\nabla_{\spoplus} = 2( \optplan{\cost} - \optplan{2 \predcost - \cost} )
\end{equation}
This subgradient is used for gradient-based training in DFL.
\section{DFL in the Context of Planning}

A classical planning problem is essentially a compact definition of a huge graph akin to a finite automaton. The objective is to find a path in this graph from a given initial state to a goal state without explicitly building the graph. In practice, we further want not only to generate \emph{some} plan, but one minimizing the costs of the contained actions. 
%
%
%
%
All existing techniques in classical planning assume that actions costs are non-negative (negative action costs constitute a very different form of problem as, in that setting, an action sequence may become cheaper when continued). This assumption presents a challenge for our DFL setting, not only because the ML model may predict costs of some actions to be negative; but also during training, there is the aspect of having to solve the planning problem with negative action costs.

\subsection{Regret Evaluation in the Presence of Negative Action Costs}
\label{sect:eval}
We highlight that while the ground truth action costs are positive, the predicted cost, returned by the ML model, might turn negative.
The explanation for why this could occur is provided in Appendix \ref{Appendix_explneg}.
One could use a Relu activation layer to enforce the predicted costs $\predcost$ to be non-negative. This can be formulated by using an element-wise $\bm{\max}$ operator.
\begin{equation}
\label{eq:thresholding}
\relu{\predcost} = \bm{\max}( \predcost ,\bm{0})  
\end{equation}
We can naively use \emph{relu}, by feeding the planning system with action costs after setting the negative ones to zero. We refer to it as \emph{thresholding}.
However, this approach may yield a subpar plan by turning all negative predictions into zeros, losing the relative ordering of negative action costs.

%
%
Next, we propose an improved method for transforming all action costs into positive values before feeding them to the planning system.
Our idea is to add a scalar value to each element of the cost vector, if any element in it is negative.
We implement this by adding the absolute value of the smallest action cost to the cost vector. 
For a given $\predcost$, it can be computed as follows:
\begin{equation}
\label{eq:cmin}
\underline{c}= \Big| \min (0, \min(\predcost) ) \Big|
\end{equation}
\noindent where $\min(\predcost)$ is the minimum value in the cost vector $\predcost$.
Eq.~\eqref{eq:cmin} ensures that if all the elements in $\predcost$ are positive, the value of $\underline{c}$ is 0.
The \sol \ obtained after this transformation, can be expressed in the following form:
\begin{equation}
\label{eq:addmin}
	\addminoptplan{ \predcost}  = \optplan{  \predcost  + \underbar{c} }
\end{equation}
\noindent where $\underbar{c} $ is defined in Eq.~\eqref{eq:cmin}.
We refer to this approach as \emph{add-min}.
We will evaluate which among these two approaches would be suitable for evaluating regret in the presence of negative action costs.

\subsection{Training in the Presence of Negative Action Costs}
The second challenge is associated with \textit{training} an ML model in the DFL paradigm. As mentioned before, DFL involves computing the planning solution with the predicted action costs during the training of the ML model. 
As the action costs might turn negative, it also requires finding a planning solution over negative action costs.
We emphasise that 
while training 
{with the} SPO method, $(2 \predcost - \cost)$ can turn negative, even if we ensure that both $\predcost$ and $\cost$ are positive.


During evaluation, as we mentioned earlier, our aim is to create a planning solution with negative action costs.
However, this objective differs during training. In the DFL paradigm, the primary focus of solving the planning problem during training is to produce a useful gradient for training. This concept is reflected in Equation \ref{eq:spo_grad};
where the gradient does not include a planning solution 
{for} the predicted action cost $\predcost$; rather, it considers a solution 
{for} $(2\predcost - \cost)$, as it yields a suitable gradient for training.

It is obvious that the thresholding or add-min approach introduced for evaluation of regret can also be used while training.
Computing the SPO subgradient using thresholding would result in the following:
\begin{equation}
\label{eq:thresholding_training}
	\nabla_{\spoplus}^{\mathit{relu}} = 2 \big( \optplan{\cost} - 
 \throptplan{2 \predcost - \cost }
 \big)
\end{equation}
On the other hand, computing the SPO subgradient 
through the add-min approach would yield the following:
\begin{equation}
	\label{eq:addmin_training}
	\nabla_{\spoplus}^{\mathit{min}^+ } =  2 \big( \optplan{\cost} -  \addminoptplan{2 \predcost - \cost}
 \big)
\end{equation}
Note that we do not have to change $\optplan{\cost}$ as it does not have any negative element in it.

\subsection{Explicit Training Penalty}
We highlight that when training the ML model using Eq.~\eqref{eq:thresholding_training} or Eq.~\eqref{eq:addmin_training}, the conversion of negative action costs to non-negative ones occurs outside the gradient computational graph. Consequently, the ML model does not receive feedback indicating the necessity of such corrective measures before computing the regret. 
This limitation motivates us to explore alternative gradient computation techniques that not only make the ML model aware of the need for such corrective actions but also has no impact when there are no negative predictions. 

We propose to add a penalty function  in the loss function if any element of the vector $ (2\predcost - \cost)$ is negative.
\begin{equation}
\label{eq:relu_penalty}
	\nspo =   \spoplus 
	+  \lambda \; \bm{1}^\top \relu{  \cost -  2\predcost} 
\end{equation}
\noindent where, $\lambda$ signifies the weight assigned to the penalty function,  
$\bm{1}$ denotes a vector of ones with the same dimension as $\cost$.
So, $\bm{1}^\top \relu{  \cost -  2\predcost}$ is the sum of all non-zero elements in $\cost -  2\predcost$.
In this formulation,
$\bm{1}^\top \relu{  \cost -  2\predcost} $ will be zero only if $2\hat{c}(a_i) < c(a_i)$ for all actions $a_i$.
In Eq.~\eqref{eq:relu_penalty}, the second term can be viewed as a regularizer that penalizes predicting $ 2\predcost < \cost$, as for such predictions we have to make the transformation of the cost vector before feeding it to the planning system.
To train with the $\nspo$ loss, we can use the subgradient $\nabla_{\spoplus}^{\mathit{relu}}$ ~\eqref{eq:thresholding_training} or $\nabla_{\spoplus}^{\mathit{min}^+ }$~\eqref{eq:addmin_training} for the \spoplus\ part; 
we will denote 
the respective loss functions as $\nspo^{\mathit{relu}}$ and $\nspo^{\mathit{min}^+ }$. 
%
%
\subsection{From Loss to Gradient Computation}
The subgradient of $\nspo$ in Eq.~\eqref{eq:relu_penalty} can be expressed in the following form:
\begin{equation}
\label{eq:relu_penalty_grad}
    \nabla_{ \nspo } = \nabla_{\spoplus} -2 \; \lambda \;  \bm{I}_{<0} (2\predcost - \cost)
\end{equation}
\noindent We use an indicator function $\bm{I}_{<0}$, which outputs a vector with elements equal to 1 for actions $a_i$ if $2\hat{c}(a_i) < c(a_i)$.
For instance, if we use $\nabla_{\spoplus}^{\mathit{min}^+ }$ as the  $\spoplus$ subgradient, $\nabla_{ \nspo }^{\mathit{min}^+ }$ takes the following form:
\begin{align}
    \nabla_{ \nspo }^{\mathit{min}^+ } &= 
    2 \big( \optplan{\cost} -  \addminoptplan{2 \predcost - \cost} \big) -2 \; \lambda \;  \bm{I}_{<0} (2\predcost - \cost)\nonumber \\
    &=  2 \bigg( \optplan{\cost} -  \Big( \addminoptplan{2 \predcost - \cost} + \lambda \;  \bm{I}_{<0}  (2\predcost - \cost)  \Big) \bigg) \nonumber \\
    \label{eq:nspo_grad}
    &= 2 \bigg( \optplan{\cost} - \negativeoptplan{2\predcost - \cost}  \bigg)
\end{align}
\noindent {where $\negativeoptplan{2\predcost - \cost}$ is defined as follows:}
\begin{equation}
	\negativeoptplan{2\predcost - \cost} =  \addminoptplan{2 \predcost - \cost} + \lambda \;  \bm{I}_{<0} (2\predcost - \cost)
\end{equation}
\noindent Using the SPO methodology, we can use Eq.~\eqref{eq:nspo_grad} as a subgradient to minimize  $\nspo^{\mathit{min}^+ }$.

The vector $\negativeoptplan{2\predcost - \cost}$ increments the count of any action $a_i$ where $2\hat{c}(a_i) < c(a_i)$ by 1 after obtaining a solution with add-min. In other words, these actions are executed once more.
In this way it penalizes for the need to correct the predicted cost by selecting any action $a_i$ for which $2\hat{c}(a_i) < c(a_i)$.
We highlight that there might be no solution to the original planning problem corresponding to the vector representation $\negativeoptplan{\predcost}$. 
Since we added actions apart from the solution returned by the planning system, there might not even be an executable permutation of the actions represented in the vector.

\paragraph{Intuitive interpretation of the subgradient.}
The motivation behind introducing the subgradient formulation \eqref{eq:nspo_grad} is that we can associate an intuitive interpretation to it.
The intuition behind the subgradient~\eqref{eq:nspo_grad} 
is that the \sol \ $\negativeoptplan{2\predcost - \cost}$ 
increases the count of action $a_i$ if $2\hat{c}(a_i) < c(a_i)$;
which makes the corresponding elements in the subgradient vector
$2( \optplan{\cost} - \negativeoptplan {2 \predcost - \cost} )$ negative. 
As the ML model is updated using the 
opposite of the subgradient, 
the corresponding action costs are increased in the next iteration.
So, in this way we incentivize the model to avoid predicting $2\hat{c}(a_i) < c(a_i)$.

\section{Scaling up DFL for Planning Problems}
As reported by \citet{mandi2023decisionfocused}, DFL comes with substantial computational costs.
This is due to the fact that DFL requires solving the planning problem with the predicted (action) costs while training the underlying ML model. 
This means that we need to solve a planning problem repeatedly, which is computationally expensive.
This computational burden poses a significant challenge in applying DFL to real-world planning problems, often resulting in long training times.
In this section, we present some strategies to tackle this crucial issue.

\subsection{Use of Planning Techniques to Expedite Training} \label{sec:planning:algs}

As DFL involves repeatedly solving the planning problem during training, one strategy to expedite training is to use planning techniques without optimality guarantees or even solutions to relaxed planning problems (as usually done when computing planning heuristics). The advantage of this is that it is easier and faster to solve. 
Although such solutions may not be identical to the optimal ones, they can still provide a useful gradient \eqref{eq:nspo_grad}.
%
Note that the gradient computation in DFL is computed across a batch of training instances. In such cases, the exact optimal solution with the predicted action costs might not be necessary to determine the direction of the gradient update. 
A non-optimal solution, \emph{reasonably close to the true solution}, often provides a good gradient direction and suffices for gradient computation.

For integer linear problems (ILPs),
\citet{MandiDSG20} observed that solving their linear relaxation is sufficient for obtaining informative DFL gradients.
%
%
In planning, we have several options towards approximating the optimal solution: we can either use planning algorithms that are bounded optimal, those without optimally guarantee, or even use solutions to relaxed planning problems.
This leads us to the following settings, where both plan quality and computational effort decrease:
\begin{itemize}
 \item $\solOpt$ -- Use an optimal planning system to get an optimal solution.
 \item $\solBound{n}$ -- Use an algorithm that guarantees a solution not worse than $n$ times the optimal plan. 
 \item $\solNonOpt$ -- Use a planning system without optimality guarantees.
 \item $\solHeuOpt{}$ -- Return a solution to a relaxation of the planning problem as usually done to compute
 \underline{h}euristics in planning. 
\end{itemize}

%

In our experiments, we use an $A^*$ search and the admissible LM-Cut heuristic~\cite{HelmertD09} for optimal planning (\solOpt).
For bounded optimal planning (\solBound{n}), we combined LM-Cut with a weighted $A^*$ search. In the latter setting, the heuristic value is multiplied with a factor, which practically leads to finding solutions more quickly, but comes at the cost of losing the guarantee of finding an optimal solution. However, solutions are guaranteed to be bounded optimal.

For planning without optimality guarantees (\solNonOpt), using a non-admissible heuristic is usually the better option to find plans more quickly. In our experiments, we combine a Greedy Best First Search (GBFS) with the \hff{} heuristic~\cite{HoffmannN01}.
\hff{} internally finds a solution to the so-called delete-relaxed (DR) planning problem, which ignores the delete effects in the original planning problem. This simplifies the problem and makes it possible to find a solution in polynomial time (while finding the \emph{optimal} solution is still NP-hard). The heuristic estimate is then the costs of the solution to the DR problem.

For the last option (\solHeuOpt{}), we need to choose a heuristic that computes not only a heuristic value, but also a relaxed plan, because we need one to compute the gradient as discussed above. Since \hff{} internally computes a DR solution, it is well suited for our setting and we can use the DR solution as well as the heuristic estimate computed by the \hff{} heuristic for our learning process.

\subsection{Use of Solution Caching to Expedite Training}\label{sec:caching}


%

As shown by \citet{MulambaMD0BG21},  an alternative approach to tackle the scalability of DFL is to replace solving an optimization problem with a \emph{cache} lookup strategy, where the cache is a set of feasible solutions and acts as an inner approximation of the convex-hull of feasible solutions. How this cache is formed is crucial to the success of this approach.
%

\citet{MulambaMD0BG21}
propose to keep all the solutions in the training data in the cache. Moreover, as the predicted action costs may deviate significantly from true action costs, particularly in early training stages, their solutions may be different from the solutions in the training instances.
To address this, they solve the 
problem for a percentage, $p\%$, of 
predicted action costs and include the corresponding solutions in the cache as well. 
Hence, this approach reduces the computational burden by a margin of $p\%$.
They report that keeping $p$ as low as $5\%$ is often sufficient for DFL training. We will implement implement this approach by caching \sol s\  and investigate whether such a solution caching approach would speed up training  without compromising the quality of decisions.
\raggedbottom

\section{Experimental Evaluation}
In this section, we first describe our benchmark set and the system setup. We come to the results afterwards.
The code and data have been made publicly available \footnote{\url{https://github.com/ML-KULeuven/DFLPredict-Action-Costs-for-Planning}}.
\subsection{Experimental Setup}
\begin{table*}[tb]
		\centering
\caption{Evaluation of learning based on optimal plans ($\solOpt$) for small-size problem instances. We report \textbf{percentage regret.} }
\label{tab:SmallPlanning}
\resizebox{\linewidth}{!}{
\begin{NiceTabular}[b]{@{}cccccccccccc}
& \multicolumn{2}{c}{Shortest Path}  &&
\multicolumn{4}{c}{Transport} &&
\multicolumn{3}{c}{Rovers} \\
 \cmidrule{2-3} \cmidrule{5-8} \cmidrule{10-12}
& SP-5 & SP-10 && 5-1-1 (a) & 5-1-1 (b) & 5-2-1 (a) & 5-2-1 (b) && Rovers1 & Rovers2 &  Rovers3 \\
\toprule
MSE & $9.37 \pm 0.17$ & $12.55 \pm 0.11$ && $9.38 \pm 0.08$ & $7.43 \pm 0.18$ & $7.74 \pm 0.07$  & $8.59 \pm 0.04$ & & $4.18 \pm 0.01$ & $4.68 \pm 0.01$ & $1.45 \pm 0.01$ \\ 
$\spoplus^{\mathit{relu}}$ & $27.82 \pm 5.21$ & $39.76 \pm 2.29$ && $16.27 \pm 0.98$ & $10.49 \pm 1.81$ & $10.99 \pm 0.84$ & $14.15 \pm 2.34$ && $7.16 \pm 0.92$ & $12.1 \pm 0.57$ & $2.21 \pm 0.23$  \\
$\spoplus^{\mathit{min}^+ }$ & $8.13 \pm 0.16$ & $9.63 \pm 0.09$ && $8.9 \pm 0.2$  &  $7.67 \pm 0.42$ & $7.72 \pm 0.44$ & $9.19 \pm 0.36$ && $5.35 \pm 0.32$ & $4.95 \pm 0.24$& $\bm{1.2 \pm 0.03}$ \\
$\nspo^{\mathit{relu}}$ & $\bm{8.07 \pm 0.09}$ & $\bm{9.16 \pm 0.03}$ & & $8.0 \pm 0.11$ & $6.03 \pm 0.18$ & $5.33 \pm 0.15$ & $\bm{6.75 \pm 0.13}$ && $\bm{3.94 \pm 0.12}$ & $4.18 \pm 0.1$ & $1.22 \pm 0.01$ \\
$\nspo^{\mathit{min}^+ }$ & $8.12 \pm 0.14$ & $9.41 \pm 0.17$ && $\bm{7.93 \pm 0.06}$ & $\bm{5.97 \pm 0.15}$ & $\bm{5.07 \pm 0.07}$ & $6.86 \pm 0.05$ && $4.02 \pm 0.1$ & $\bm{4.13 \pm 0.09}$ & $1.21 \pm 0.04$ \\
\bottomrule
\end{NiceTabular}
   }
\end{table*}
\subsubsection{Benchmark Set}
For our experiments we need domains with meaningful action costs that have impact on solution quality (otherwise we will not be able to measure the impact of our methods). Further, to have a wide range of solving techniques available we want to stay in the standard classical planning (i.e., non-temporal) setting.  
We use a problem generator to generate problems of different sizes.
In the Rovers domain, meeting these requirements required some adjustments. Next, we detail the domains, their source, and (if necessary) modifications we made. 

%


\paragraph{Shortest path.}
This domain models a $n\times n$ grid environment an agent needs to navigate through.
Each node is connected to its top and right nodes.
The objective is to find a path starting from the bottom left cell to the top right cell with minimal costs. 
This domain is particularly interesting for our experiments, because it is a widely used benchmark in DFL \citep{spo, mandi2023decisionfocused, tang2022pyepo}. In these works, the problem is solved using an LP solver. We include this to have a direct comparison to existing DFL methods.
\paragraph{Transport.}
In this domain we use the standard domain and generator~\cite{seipp-et-al-zenodo2022} from the generator repository\footnote{https://github.com/AI-Planning/pddl-generators}.
Each transport problem instance revolves around the task of delivering $p$ number of packages using $t$ number of trucks. 
We consider a $n \times n$ grid for the transport problem, within which both pickup and delivery operations occur.
We denote each transport problem instance as $n$-$p$-$t$, signifying that the grid is of $n\times n$ dimension, with p representing the number of packages and t indicating the available truck count. 
\paragraph{Rovers.} 
This domain describes the task of a fleet of Mars rovers, each equipped with a (probably different) set of sensors. They need to navigate and gather data (e.g.\ rock samples or pictures). The data then needs to be send to the lander.
%
%
%
%
Our domain is based on the one from the 2006 International Planning Competition.
However, the domains from the different competition tracks did not directly fit our needs:
\emph{MetricTime} contains durative actions,
\emph{Propositional} and \emph{QualitativePreferences} do not include action costs.
We created a model based on the \emph{MetricSimplePreferences} track and made the preferences normal goals. To get integer costs, we multiplied the included action costs by $10$ and rounded them afterwards to integers.



%
For our domains, we generated two groups of problem instances: small-sized instances that can be solved within 0.25 seconds, and large-sized instances that take 0.5–1 seconds to solve.
For more details, please refer to the Appendix \ref{solving_time}.
\begin{table}[hb]
		\centering
\caption{Evaluation on Shortest path problem instances trained using LP solver with and without \emph{relu}. We report \textbf{percentage regret.}}
\label{tab:LP_relu}
\begin{tabular}{@{}ccccc@{}}
\toprule
 & Without Relu &  With Relu \\
  \toprule
    \multicolumn{3}{c}{SP-5} \\
    \midrule
    MSE & $\bm{9.37 \pm 0.17}$ & $13.8 \pm 0.61$ \\
    \spoplus & $\bm{8.26 \pm 0.15}$   & $10.89 \pm 0.42$ \\
    \midrule
    \multicolumn{3}{c}{SP-10} \\
    \midrule
    MSE & $\bm{12.55 \pm 0.11}$ & $15.79 \pm 0.17$ \\
    \spoplus & $\bm{9.44 \pm 0.13}$ & $12.89 \pm 0.57$ \\
\bottomrule
\end{tabular}
\end{table}
\subsubsection{Generation of Training Data}
While we adopt the planning problems from planning benchmark domains, 
we synthetically generate the action costs. Such synthetic data generation processes are common in DFL literature. We follow the synthetic data generation process exactly as done by \citet{spo}.
%
We generate a set of pairs of features and action costs $\{ (\features_{\kappa}, \cost_{\kappa}) \}_{\kappa=1}^N$ for training and evaluation.
The dimension of $\cost_{\kappa}$ is equal to the number of actions, $ |\plActions|$, which is specific to the planning problem.
The dimension of $\features_{\kappa}$ is 5.
and each $\features_{\kappa}$ is sampled from a multivariate Gaussian distribution with zero mean and unit variance, i.e., $\features_{\kappa} \sim \mathbf{N} (0, I_5)$ ($I_5$ is a $5 \times 5$ identity matrix).
To set up a mapping from $\features_{\kappa}$ to $\cost_{\kappa}$, first, a matrix $B \in \mathbb{R}^{|\plActions| \times 5}$ is constructed, and then $\cost_{\kappa}$ is generated according to the following formula:
\begin{equation}
    c^{\kappa } (a_i) = \bigg [ \bigg(\frac{1}{\sqrt{p}} \big(B \features_{\kappa}  \big) +3  \bigg)^{\textit{Deg}  } +1 \bigg]\xi_{\kappa}^i
\end{equation}
where $c^{\kappa } (a_i)$ is the the cost of action $i$ in instance $\kappa$, the parameter \emph{Deg} parameter signifies the extent of \emph{model misspecification}, and $\xi_{\kappa}^i$ is a multiplicative noise term sampled randomly from the uniform distribution. 
Note that the action costs generated in this manner are always \emph{positive} if \emph{Deg} is a even number. Furthermore, since the action costs are random numbers sampled from a continuous distribution, it is highly improbable that two feasible plans will have exactly identical execution costs. Therefore, in this scenario, we do \emph{not} encounter the phenomenon of multiple non-unique solutions.

\citet{spo} use a \emph{linear model} to predict the cost vector from features.
The higher the value of \emph{Deg}, the more the true relation between the features and action costs deviates from the linear model and the larger the errors of the linear predictive model.
Such model misspecification is a common phenomenon in ML, because the in practise the data generation process is not observable.
In our experiments, we will report result with \emph{Deg} being 4.

\subsubsection{Planning and Learning Setup}
Similarly to \citeauthor{spo}~\citep{spo}, we will also use a linear model to predict the action costs from features.
We use PyTorch~\citep{paszke2017automatic} to implement the linear predictive model and train it by minibatch stochastic gradient descent \citep{goyal2017accurate, robbins1951stochastic}. 
The gradient is backpropagated for training the model using PyTorch's automatic differentiation.
%
%
As a planning tool, we use the Fast Downward (FD) planning system~\cite{Helmert06} and run the algorithms described in Section~\ref{sec:planning:algs}.
For small-size planning problems, we generated 400, 100 and 400 training, validation and test instances.
For large-size planning problems, these values are 200, 25 and 25 respectively.
%

\begin{table}[hb]
		\centering
\caption{Comparison between \emph{add-min} and \emph{thresholding} regret of models trained using LP solver for the shortest path solver. We report their \textbf{deviations from the regret evaluated using an LP solver}.}
\label{tab:LP}
\begin{tabular}{@{}ccccc@{}}
\toprule
  &  \makecell{Thresholding  \\Difference} & \makecell{Add-min\\Difference}    \\
  \toprule
     \multicolumn{3}{c}{SP-5} \\
    \toprule
    MSE &  $0.38 \pm 0.32$ & $\bm{0.0 \pm 0.01}$ \\
    \spoplus &  $10.55 \pm 1.18$ & $\bm{0.0 \pm 0.01}$ \\
    \bottomrule
     \multicolumn{3}{c}{SP-10} \\
    \toprule
    MSE & $-0.74 \pm 0.17$ & $\bm{0.02 \pm 0.04}$ \\
    \spoplus &  $14.68 \pm 1.65$ & $\bm{0.01 \pm 0.02}$  \\
\bottomrule
\end{tabular}
\end{table}
\subsection{Results}
\begin{table*}[h!]
	\centering
	\caption{Evaluation of models trained with different planning techniques without optimality guarantees with $\nspo^{\mathit{min}^+ }$ for large-size problem instances.  We report \textbf{percentage regret} and  \textbf{training time of 20 epochs in seconds}. We highlight those which have lower regret than MSE.
 }
	\label{tab:heur}
	 \resizebox{\linewidth}{!}{
		\begin{NiceTabular}{@{}ccc|cccc@{}}
			\toprule
                && & $\solOpt$ & $\solBound{n}$ & $\solNonOpt$ & $\solHeuOpt{}$ \\
			Problem&  &MSE & $\mathit{A^*}$  with LM-Cut & $\mathit{WA^*(2)}$ with LM-Cut& GBFS with \hff{}  & \hff{} del. relaxed plan \\
			\toprule
   \multicolumn{8}{c}{\textbf{Transport Problem}}\\
   \bottomrule
  \multirow{ 2}{*}{5-3-1} & Regret & $5.84 \pm 0.26$ & ${ \bm{4.19 \pm 0.4 }}$ & $6.06 \pm 0.8$ & $9.2 \pm 0.9$ & $8.06 \pm 0.41$   \\
  & Training Time & 350 & 4800 & 1700  & 700 & 250\\
  \midrule
    \multirow{ 2}{*}{5-2-2} & Regret & $14.15 \pm 0.0$ & ${ \bm{11.4 \pm 0.89}}$ & ${ \bm{12.43 \pm 0.73}}$ & ${ \bm{13.12 \pm 1.23}}$ &  ${ \bm{ 13.1 \pm 1.01 }}$ \\
  & Training Time & 350 & 9050 & 5050  & 800 & 200\\
  \midrule
    \multirow{ 2}{*}{10-1-1} & Regret 
   & $12.99 \pm 0.17$  & ${ \bm{12.16 \pm 0.8} }$  & ${ \bm{12.55 \pm 1.34}}$ & $16.86 \pm 1.39$ & $15.65 \pm 0.89$ \\
   & Training Time & 100 & 3650 & 3550  & 700 & 100\\
  \toprule
  \multicolumn{8}{c}{\textbf{Rovers Problem}}\\
   \bottomrule
  \multirow{ 2}{*}{Rovers4} & Regret  &$2.69 \pm 0.05$ & ${ \bm{2.3 \pm 0.15}}$ & $2.78 \pm 0.15$  & $3.66 \pm 0.49$ & $4.97 \pm 0.27$\\
   & Training Time & 250 & 9300 & 1550  & 700 & 200\\
   \midrule
   \multirow{ 2}{*}{Rovers5} & Regret &  $2.92 \pm 0.09$  & ${ \bm{2.91 \pm 0.25}}$ & $3.8 \pm 0.73$ & $5.36 \pm 0.44$ & $5.76 \pm 0.21$ \\
    & Training Time & 300 & 10300 & 850  & 700 & 200\\
			\bottomrule
		\end{NiceTabular}
		 }
\end{table*}
%
In this section, we will present key insights from our empirical evaluation. 
After training the ML model, we report \emph{percentage regret} on the test data, which is computed as follows:
\begin{equation}
\label{eq:relative_regret}
\frac{1}{N_{test}} \sum_{\kappa=1}^{N_{test}} \frac{ \cost_{\kappa}^\top \optplan{\predcost_{\kappa}} - \cost_{\kappa}^\top \optplan{\cost_{\kappa}} } { \cost_{\kappa}^\top \optplan{\cost_{\kappa}}}.
\end{equation}
For each set of experiments, we run 5 experiments each time with different seeds and report average and standard deviation of 
percentage regret in the tables.
We confirmed that all the models converge within 20 epochs. We report results after the $20^{\text{th}}$ epoch.

\subsubsection{Evaluating Regret for Planning Problems}
\paragraph{RQ1: Does training with a \emph{relu} activation layer impact regret?}
One can enforce the predictions to be non-negative by adding \emph{relu} as a final activation layer.
However, when the predictive model does not fully represent the data generation process,
imposing non-negativity constraint using \emph{relu} may distort the predictions resulting in higher prediction as well as decision errors.
To investigate whether using \emph{relu} affects the regret, we consider the Shortest path problem, which is a widely used benchmark in DFL. 

As this is a shortest path problem over a directed \emph{acyclic} graph, negative action costs cannot lead to loops and degenerate behaviour.
Hence, we can obtain the true \emph{optimal} solution even in the presence of negative action costs using an LP solver.
In this experiment, for both training and evaluation, we use Gurobi LP solver \citep{gurobi}.
We observe in Table \ref{tab:LP_relu} that for both MSE and $\spoplus$,
the regret increases as we use \emph{relu} activation layer. 
From this we conclude, we are better-off without the \emph{relu} activation layer.

\paragraph{RQ2: How to evaluate regret given that planning system does not allow negative costs?}
As we will not be using \emph{relu} activation layer in the final layer, the predictions generated by the ML model can turn negative, even though
the groundtruth action costs are positive.  
As action costs with negative values, are not supported by a planner; we will be using \emph{thresholding} \eqref{eq:thresholding} or \emph{add-min} \eqref{eq:addmin} to solve the planning problem with negative predicted action costs. 
We again consider the shortest path problem. This time we again use the LP solver for training. However, during evaluation, we compute regret using both the LP solver and a planner, allowing to compare the true regret with the regret obtained by a planner.
We want to find out which method, \emph{thresholding} or \emph{add-min}, gives a regret measure closest to the LP regret.
We see in Table~\ref{tab:LP} that
\emph{add-min} regret demonstrates greater fidelity to true LP regret. Note that \emph{thresholding} regret shows significant deviations, particularly evident $\spoplus$ .
Hence in our latter experiments, \textbf{we will use \emph{add-min} regret} to evaluate the regret of predicted action costs. With the evaluation protocol set, we now focus on DFL learning methods.
\begin{table*}[bt!]
	\centering
	\caption{Evaluation of $\nspo^{\mathit{min}^+ }$ trained with optimal plans ($\solOpt$) and caching $p= 10\%$ and $20\%$ for large-size problem instances. We report \textbf{percentage regret} and  \textbf{training time of 20 epochs in seconds}. We highlight those which have lower regret than MSE.}
	\label{tab:caching}
	 \resizebox{\linewidth}{!}{
		\begin{tabular}{@{}ccccccccccccc@{}}
			\toprule
			& \multicolumn{2}{c}{MSE} &  & \multicolumn{2}{c}{$\solOpt$}  & & \multicolumn{2}{c}{Caching($p=10\%$)} & & \multicolumn{2}{c}{Caching($p=20\%$)}\\
   \cmidrule{2-3} \cmidrule{5-6} \cmidrule{8-9} \cmidrule{11-12}
   Problem & Regret &  Time & &  Regret & Time & &   Regret & Training Time & &   Regret &  Time \\
			\toprule
   \multicolumn{12}{c}{\textbf{Transport Problem}}\\
   \bottomrule
  5-3-1 & $5.84 \pm 0.26$ & 350 && ${ \bm{4.19 \pm 0.4}}$ & 4800 & & $5.85 \pm 0.75$ & 800 & & ${ \bm{4.7 \pm 0.52}}$ & 1050 \\
    5-2-2 & $14.15 \pm 0.0$ & 350 && ${ \bm{11.4 \pm 0.89}}$ & 9050 && ${ \bm{11.03 \pm 1.57}}$  & 900 && ${ \bm{11.07 \pm 1.31}}$ & 1550 \\
    10-1-1 & $12.99 \pm 0.17$ & 100 && ${ \bm{12.16 \pm 0.8}}$ & 3650 &&  $14.5 \pm 1.28$ & 450 && ${ \bm{12.07 \pm 1.1}}$ & 800\\
  \toprule
  \multicolumn{12}{c}{\textbf{Rovers Problem}}\\
   \bottomrule
  Rovers4 & $2.69 \pm 0.05$ & 250 &&  ${ \bm{2.3 \pm 0.15}}$ & 9300 & & $2.72 \pm 0.34$ & 1050 &&  ${ \bm{2.29 \pm 0.22}}$ & 2000 \\
   Rovers5 & $2.95 \pm 0.0$ & 300 && ${ \bm{2.81 \pm 0.05}}$ & 10300 & &  $3.55 \pm 0.18$ & 1250 &&   ${ \bm{2.92 \pm 0.38}}$ & 2300 \\
			\bottomrule
		\end{tabular}
		 }
\end{table*}
\subsubsection{Training With and Without Explicit Penalty}
\paragraph{RQ3: How do the proposed SPO subgradients perform?}


After comparing with related work on the \emph{Shortest path} domain, we evaluate our methods on the \emph{Transport} and \emph{Rovers} domain known from the planning literature.
So, in this case, we use the FD planner for DFL training as well as evaluation.
We seek to answer whether adding the explicit training penalty results in lower regret.
As DFL training requires repeatedly solving a planning problem for every training instance,
we restrict ourselves to planning problems that are fast to solve.
We consider small-size planning problems, which can be solved quite fast (within 0.25 seconds).
In an earlier stage, we experimented with different integer $\lambda$ values and found that $\lambda =1$ resulted in the lowest regret. A higher $\lambda$ increases the influence of the penalty in the final loss \eqref{eq:relu_penalty}, reducing the impact of SPO+ loss. 

We report the result in Table \ref{tab:SmallPlanning}. 
$\spoplus^{\mathit{relu}}$ loss performs very poorly, as its regret is much higher than MSE. 
This is due to the fact that turning negative costs to zero without considering their values causes loss of information.
$\spoplus^{\mathit{min}^+ }$ performs much better. However, even in some cases its regret is higher than MSE.
On the other hand, $\nspo^{\mathit{relu}}$ and $\nspo^{\mathit{min}^+ }$, which add explicit training penalty if 
$2\hat{c}(a_i) < c(a_i)$ for a action $a_i$, are able to improve $\spoplus^{\mathit{relu}}$ and $\spoplus^{\mathit{min}^+ }$.
It is interesting to note that the difference between $\mathit{relu}$ and $\mathit{min}^+$ is insignificant after adding explicit training penalty.
This experiment suggests both $\spoplus^{\mathit{relu}}$ and $\spoplus^{\mathit{min}^+ }$ are effective DFL approaches for predicting action costs in planning problems.

\subsubsection{Optimal Planning Versus Non-Optimal Planning}
\paragraph{RQ4: Can we use non-optimal planning for DFL training?}
As DFL requires solving the planning problem repeatedly while training, which creates a considerable computational burden
when challenging planning problems are considered.
Hence we seek to answer whether we can utilize non-optimal planning algorithms in DFL.

To investigate this, we consider larger problem instances (solving such instance requires between 0.5 and 1.5 seconds).
We train each time with $\nspo^{\mathit{min}^+ }$ loss; but non-optimal planning and plans for DR planning problems.
In Table~\ref{tab:heur}, we observe that DFL training with 
$\solNonOpt$ and $\solHeuOpt{}$
results in considerably higher regret. 
The regret of $\solBound{n}$ is higher than $\solOpt$, but mostly lower than $\solNonOpt$ and $\solHeuOpt{}$.
However, in most cases its regret is higher than MSE regret, which is not desirable.

\subsubsection{Optimal Planning versus Solution Caching}
\paragraph{RQ5: Can we use solution caching to speed up training?}
Next we investigate whether solution caching, as implemented by \citet{MulambaMD0BG21} in the context of DFL for optimization, is effective for planning problems too. 
We initialize the cache with all the solutions present in the training data.
We 
experiment with $p= 10\%$ and $20\%$.
%
We can see in Table~\ref{tab:caching}, the training time of caching faster compared to  $\solOpt$ due to they solve the planning is solved for only $p\%$ of instances using the predicted action costs.
While $p=10\%$ does not consistently outperform MSE regret, $p=20\%$  produces regret lower than MSE for all instances.
This indicates for large planning instances, use of solution caching with $p=20\%$ could prove to be a useful approach.
\section{Conclusion}


In this work, we investigated for the first time how we can use techniques from DFL in the planning domain. More specifically for the case of predicting action costs from correlated features and historic data, we showed how the SPO technique, which places no assumptions on the solver used, can also be used for planning problems. Other DFL techniques which work with a black-box solver~\cite{ PogancicPMMR20,niepert2021implicit} are now equally applicable.

We proposed an implementation of DFL which accounts for the fact that planners do not support negative action costs. Our findings suggest that imposing non-negativity through relu leads to an increase in regret, both for model trained with MSE and DFL loss.
Moreover, training with an explicit penalty for correcting negative action costs before solving the planning problem yields significant improvements. Indeed, our DFL approach always leads to lower regret than when training to minimize MSE of predicted action costs. While using sub-optimal plans did not consistently lead to lower-than-MSE regret, a moderate amount of caching was able to reduce computation cost significantly.

%
Future work includes reducing computational costs further, as well as DFL for state-dependent action cost prediction or other action components; for which SPO and related techniques is insufficient.

\begin{ack}
This research received funding from the European Union’s Horizon 2020 research and innovation programme under grant agreement
No 101070149, project Tuples.
Jayanta Mandi is supported by the Research Foundation Flanders (FWO) project G0G3220N.
\end{ack}
\bibliography{literature}
\bibliographystyle{plainnat}
\setcounter{table}{0}
\renewcommand{\thetable}{A\arabic{table}}
\setcounter{figure}{0}
\renewcommand{\thefigure}{A\arabic{figure}}

\appendix
\section{Appendices}
\subsection{Details of Transport Problem Instances}
\begin{center}
\begin{table}[h]
 \caption{Specification of Transport Problem Instances}
    \centering
    \begin{tabular}{ccccc}
    \toprule
     & \makecell{Pickup Location\\ of package(s)}& \makecell{Drop Location\\ of package(s)}  & Location of Truck(s) \\
     \midrule
        5-1-1(a) & (5,5) & (1,1) & (1,3)\\
        \midrule
        5-1-1(b)  & (2,2) & (1,1) & (1,3)\\
         \midrule
        5-2-1(a) & \makecell{(5,5)\\ (1,5)} &
        \makecell{(2,1)\\ (5,1)} & (3,3)\\
         \midrule
        5-2-1(b) & \makecell{(5,4)\\ (4,3)} & \makecell{(2,1)\\ (1,1)} & (5,5)\\
         \midrule
       5-3-1 &  \makecell{(1,4)\\ (4,3) \\(3,4)} &  \makecell{(5,4)\\ (5,1) \\(4,1)} & (1,1)\\
        \midrule
        5-2-2 &  \makecell{(1,4)\\ (1,5)} &  \makecell{(5,4)\\ (3,3)}  &  \makecell{(1,1)\\ (5,5)}\\
         \midrule
        10-1-1 & (3,9) & (1,1) & (10,1) \\
         \bottomrule
    \end{tabular}
   
    \label{tab:my_label}
\end{table}
\end{center}

\subsection{Details of the Rovers Problem Instances}
\begin{center}
\begin{table}[h]
 \caption{Specification of Rovers Problem Instances}
    \centering
    \begin{tabular}{ccccc}
    \toprule
     & Number of Rovers & Number of Waypoints & Number of Camera & Number of Goals \\
     \midrule
        Rovers1 & 1 & 7 & 1 & 3\\
        \midrule
        Rovers2 & 1 & 8 & 1 & 4\\
         \midrule
        Rovers3 &  1 & 10 & 1 & 5\\
         \midrule
        Rovers4 &  1 & 10 & 3 & 6\\
         \midrule
        Rovers5 &  1 &  10 & 3 & 6\\
         \bottomrule
    \end{tabular}
   
    \label{tab:my_label}
\end{table}
\end{center}
\clearpage

\subsection{Explanation of Negative Predicted Action Costs}
\label{Appendix_explneg}
Negative predictions of action costs may occur when the predictive model is misspecified.
 For instance, if the actual relationship between $\cost$ and $\features$ is $\cost =2 \features^2 - 4 \features+ 3$ and we fit a linear model like $\predcost = \alpha  \features+ \beta$, the predicted model might have $\alpha =10$ and $\beta= -15$. Consequently, for $\features < 1.5$, the predicted $\predcost$ is negative.
It might also happen due to very high value outliers.
An example of negative predictions due to high positive outlier values: Suppose the true model is $\cost = 2 \features + 2$ but $\cost$ values corresponding to high $\features$ are affected by high noise. This might turn the predicted slope to be steeper, e.g., $\predcost =  4 \features - 1$. Consequently, for $\features < 0.25$, $\predcost <0$. 

\begin{figure*}
    \centering
     \begin{subfigure}[t]{0.45\textwidth}\includegraphics[scale=0.3]{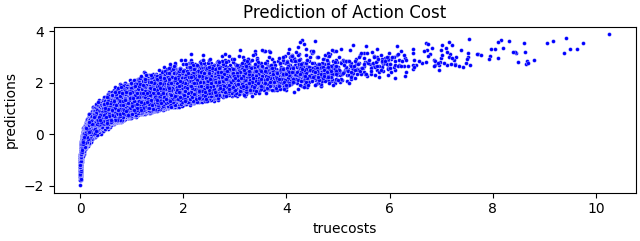}
     \caption{Prediction without Relu activation}
     \label{fig:norelu}
    \end{subfigure} 
     \begin{subfigure}[t]{0.45\textwidth}\includegraphics[scale=0.3]{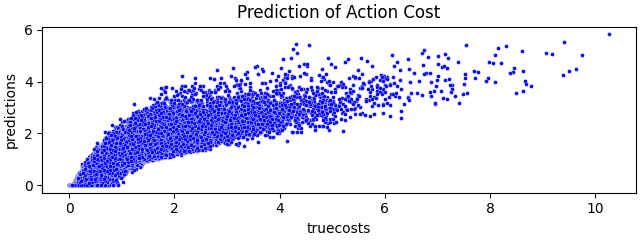}
     \caption{Prediction with Relu activation}
     \label{fig:relu}
    \end{subfigure}
    \vskip 10pt

    \caption{Comparison between prediction with or without Relu. Relu layer generates prediction chopping of negative action costs. This is why the quality of the solutions using relu layer deteriorates. }
    \label{fig:relu-pred}
\end{figure*}

In Figure \ref{fig:norelu}, we empirically demonstrate why there might be negative predicted action cost even if all the true action costs are positive. 
For this demonstration, the data is generated synthetically following the procedure of the SPO paper \cite{spo}.
The value of the \textit{Deg} parameter is 4, so the true relationship between $\cost$ and $\features$ is non-linear. However, a linear model is fit on the dataset by minimizing mean square error loss.  As shown in the numerical example, the predicted $\predcost$ becomes negative when $\cost$ is low.
Figure \ref{fig:relu} presents a scatterplot comparing predicted action costs to true action costs when the predictive model uses \textit{ReLU} activation as the final layer.
The prediction quality is poor, as all negative $\predcost$ values are indiscriminately set to zero, ignoring the magnitude of the negative values in absolute terms.

\subsection{Solving Time}
\label{solving_time}
\begin{table*}[htb]
		\centering
\caption{Solution Time }
\label{tab:SmallPlanning}
\begin{NiceTabular}{@{}ccccc@{}}
\toprule
Name &  Time (sec.)  \\

    \toprule
    \multicolumn{6}{c}{\textbf{Shortest Path Problem}}\\
SP-5 & 0.03\\
SP-10 & 0.03\\
    \toprule
    \multicolumn{6}{c}{\textbf{Transport Problem}}\\
5-1-1(a) &  0.06 \\
5-1-1(b)  & 0.11 \\
5-2-1(a) & 0.15 \\
5-2-1(b) & 0.1\\
5-3-1 &   0.25\\
5-2-2 &   1.98\\
10-1-1&   0.67 \\
    \multicolumn{6}{c}{\textbf{Rovers Problem}}\\
    \toprule
Rovers1 & 0.05\\
Rovers2 & 0.09\\
Rovers3 & 0.15\\
Rovers4 & 1.50\\
Rovers5 & 1.80\\
\bottomrule
\end{NiceTabular}
\end{table*}

\medskip

\end{document}